\documentclass[11pt]{article}
\pdfoutput=1
\usepackage[margin=1.0in]{geometry}
\usepackage{amsmath,amssymb}
\usepackage{booktabs}
\usepackage{graphicx}
\usepackage{hyperref}
\usepackage{microtype}
\usepackage{xcolor}

\hypersetup{
  colorlinks=true,
  linkcolor=blue!60!black,
  citecolor=blue!60!black,
  urlcolor=blue!60!black
}

\title{\textbf{Auditing and Adapting kNNGuard for Long-Context Prompts}}
\author{Niccol\`{o} Ronchetti}
\date{}

\begin{document}
\maketitle

\begin{abstract}
Training-free activation guardrails such as \textsc{kNNGuard}~\cite{abdelfattah2026knnguard} score prompts by comparing last-token LLM hidden states (the Layer Ensemble branch, \textsc{kNNGuard LE}) and sentence embeddings (\texttt{all-MiniLM-L6-v2}, \textsc{Embedding-kNN}) against a small reference bank ($n=50$ examples per class), combining the two branches into a Fused Ensemble (\textsc{kNNGuard FE}).

While effective on short inputs ($\le 256$ tokens), standard \textsc{kNNGuard} degrades substantially on long inputs ($>1000$ tokens): averaged across \texttt{Qwen3.5-4B} and \texttt{Gemma-4-E4B-it}, \textsc{kNNGuard LE} AUROC drops from $0.742$ ($0.777$ and $0.706$) to $0.554$ ($0.587$ and $0.521$) on Content Harm Moderation (\texttt{Aegis-2.0}), and from $0.994$ ($0.995$ and $0.992$) to $0.556$ ($0.598$ and $0.513$) on RAG Hallucination Detection (\texttt{HaluBench}).\\
We trace this degradation to three causes: (1) hidden-state drift as sequence length increases~\cite{kale2026promptinjection}, (2) hard 256-token truncation in \texttt{MiniLM}, and (3) signal dilution when a long document is compressed into a single final-token vector.
We then evaluate three training-free adaptations: a \textbf{Length-Routed Dual Bank}, a long-context embedding model (\texttt{Qwen3-Embedding-0.6B}), and \textbf{Strided Window Pooling} across intermediate token positions.\\
On Content Harm Moderation, dual-bank routing and strided window pooling raise long-context \textsc{kNNGuard LE} AUROC from $0.554$ to $\mathbf{0.828}$ ($+0.274$ AUROC) and fused \textsc{kNNGuard FE} AUROC from $0.721$ to $\mathbf{0.886}$ ($+0.164$ AUROC) with no latency overhead from bank routing.
On RAG Hallucination Detection, where the guardrail must verify logical consistency between a generated answer and a retrieved passage, sentence-embedding $k$NN remains near random chance across all lengths, while length-matched LLM hidden states (\textsc{kNNGuard LE}) recover medium-length detection ($\text{AUROC} = 0.853$) but remain limited on $>1000$-token passages ($\text{AUROC} = 0.607$) due to final-token dilution over long contexts.
\end{abstract}

\section{Introduction}
As Large Language Models (LLMs) are deployed in retrieval-augmented generation (RAG)~\cite{lewis2020retrievalaugmented}, long-document analysis~\cite{bai2023longbench}, and multi-paragraph moderation pipelines~\cite{rebedea2023nemo}, guardrails must screen inputs and outputs spanning thousands of tokens without adding high latency or retraining costs.

Most production guardrails are separate open-weight LLMs (typically 7B to 12B parameters) that have been fine-tuned on labeled safety or hallucination datasets to act as classifiers.
Recent examples include \textsc{Llama Guard 4}~\cite{metallama2025llamaguard4}, \textsc{Llama Nemotron Safety Guard V2}~\cite{ghosh2025aegis20}, \textsc{Llama Nemotron Topic Guard}~\cite{sreedhar2024canttalkaboutthis}, \textsc{Qwen3Guard}~\cite{zhao2025qwen3guard}, \textsc{Granite Guardian}~\cite{padhi2024granite}, and \textsc{Lynx}~\cite{ravi2024lynx}.
While accurate on their training distributions, running a dedicated classifier LLM adds inference latency to every request, and changing a safety policy requires collecting new training data and fine-tuning the model again.

To avoid fine-tuning a separate classifier, Abdelfattah et al.~\cite{abdelfattah2026knnguard} recently introduced \textsc{kNNGuard}, a training-free guardrail based on nearest-neighbor retrieval in activation space~\cite{khandelwal2019generalization} and hidden-state probing~\cite{zou2023representation,azaria2023the,marks2023the}.
For each task, \textsc{kNNGuard} caches representations for a small reference bank of $50$ negative (allowed) and $50$ positive (rejected) prompts.

At inference time, it classifies a new input in a single forward pass by combining two nearest-neighbor risk scores: one computed from the LLM's last-token hidden states across multiple transformer layers, and one computed from a lightweight sentence encoder (\texttt{all-MiniLM-L6-v2}~\cite{reimers2019sentencebert,wang2020minilm}), selecting whichever branch is more confident.

Despite strong performance on standard benchmarks, \textsc{kNNGuard}'s design implicitly assumes short inputs ($\le 256$ tokens).
In this work, we audit how \textsc{kNNGuard} behaves as input length scales from under $256$ tokens to over $6400$ tokens across two benchmarks: Content Harm Moderation (\texttt{nvidia/Aegis-AI-Content-Safety-Dataset-2.0}~\cite{ghosh2025aegis20}) and RAG Hallucination Detection (\texttt{PatronusAI/HaluBench}~\cite{ravi2024lynx}), evaluated on \texttt{Qwen/Qwen3.5-4B}~\cite{yang2025qwen3} and \texttt{google/gemma-4-E4B-it}~\cite{team2025gemma}.

We show that standard \textsc{kNNGuard} degrades on long inputs for three concrete reasons:
\begin{enumerate}
  \item \textbf{Reference-Bank Length Mismatch:} Kale and Vishwanath~\cite{kale2026promptinjection} recently showed that the distribution of last-token hidden states shifts systematically as sequence length $T$ increases, because self-attention layers aggregate over a growing number of context tokens.
  When the reference bank contains only short prompts ($\le 256$ tokens), long test queries ($>1000$ tokens) land in a shifted region of activation space, causing the LLM activation branch (\textsc{kNNGuard LE}) AUROC to drop toward random chance ($0.513$--$0.598$).

  \item \textbf{Hard Truncation and Limits of Sentence Embeddings:} The default \texttt{all-MiniLM-L6-v2}~\cite{wang2020minilm} encoder truncates inputs at $256$ tokens.
  In RAG hallucination detection (\texttt{HaluBench}), where each input places the retrieved passage before the question and the answer to be verified, a $256$-token limit cuts off the question and answer on over $60\%$ of medium-length inputs ($257$--$1000$ tokens) and $100\%$ of long inputs ($>1000$ tokens).
  Moreover, as we show in Section~\ref{sec:results}, even when truncation is completely removed using a long-context embedder or $256$-token chunking, sentence-embedding $k$NN remains near random chance ($\approx 0.46$--$0.56$ AUROC) on RAG hallucination detection because logical consistency between an answer and a passage is too fine-grained to be captured by static text embeddings.
  \item \textbf{Single-Vector Signal Dilution:} Compressing a multi-thousand-token document into a single final-token vector $\mathbf{h}^{(\ell)}[-1]$ dilutes localized unsafe statements or single-sentence factual contradictions buried inside a mostly benign passage~\cite{liu2023lost}.
\end{enumerate}

\subsection*{Our Contributions}
To isolate the effect of each failure mode and adapt \textsc{kNNGuard} to long contexts without fine-tuning, we make the following contributions:
\begin{itemize}
  \item \textbf{Long-Context Audit of \textsc{kNNGuard}:} Using the Area Under the ROC Curve (AUROC), we evaluate standard \textsc{kNNGuard} across short ($\le 256$ tokens), medium ($257$--$1000$ tokens), and long ($>1000$ tokens) inputs on \texttt{Aegis-2.0} and \texttt{HaluBench}, showing that \textsc{kNNGuard LE} AUROC falls from $0.706$--$0.777$ to $0.521$--$0.587$ on content moderation and from $0.992$--$0.995$ to $0.513$--$0.598$ on RAG hallucination detection.
  \item \textbf{Length-Routed Dual Reference Bank:} We introduce a dual-bank routing mechanism that switches between a short reference bank ($\le 256$ tokens) and a long reference bank ($>256$ tokens) based on the input token count $T$ with no latency overhead.
  On long content moderation queries (\texttt{Aegis-2.0}), dual-bank routing lifts \textsc{kNNGuard LE} AUROC from $0.554$ to $0.722$ and fused \textsc{kNNGuard FE} AUROC from $0.721$ to $\mathbf{0.886}$, while keeping short-prompt accuracy unchanged.
  \item \textbf{Long-Context Embedding Replacement:} We replace \texttt{MiniLM} with \texttt{Qwen/Qwen3-Embedding-0.6B}~\cite{zhang2025qwen3embedding} (which supports up to $32768$ tokens via Rotary Positional Embeddings~\cite{su2021roformer}), eliminating answer truncation on long inputs and improving short-context safety AUROC on \texttt{Gemma-4-E4B-it} from $0.746$ to $0.790$.
  \item \textbf{Strided Window Pooling:} Instead of relying only on a single end-of-sequence vector for a long prompt, we score the prompt across $256$-token windows (reading out the LLM hidden state at every $256$-th token during the single prefill pass for \textsc{kNNGuard LE}, and encoding $256$-token text chunks for the embedding branch) and aggregate window scores via \texttt{Top-2 Mean} pooling.
  This boosts long-context \textsc{kNNGuard LE} AUROC on \texttt{Aegis-2.0} from $0.722$ to $\mathbf{0.828}$ ($0.837$ on \texttt{Qwen3.5-4B} and $0.818$ on \texttt{Gemma-4-E4B-it}).
  Finally, we show why reference-bank adaptation recovers medium-length RAG hallucination detection ($\text{LE AUROC} = 0.853$, $\text{FE AUROC} = 0.837$) while sentence-embedding $k$NN remains near random chance ($\text{AUROC} \approx 0.46$--$0.56$) across all context lengths.
\end{itemize}

\section{Related Work and Preliminaries}
\subsection{Related Work}
\subsubsection*{Fine-Tuned Guardrail Classifiers}
Programmable guardrail toolkits such as \textsc{NeMo Guardrails}~\cite{rebedea2023nemo} and fine-tuned safety models (\textsc{Llama Guard 4}~\cite{metallama2025llamaguard4}, \textsc{Llama Nemotron Safety Guard V2}~\cite{ghosh2025aegis20}, \textsc{Llama Nemotron Topic Guard}~\cite{sreedhar2024canttalkaboutthis}, \textsc{Granite Guardian}~\cite{padhi2024granite}, and \textsc{Qwen3Guard}~\cite{zhao2025qwen3guard}) treat content moderation as a supervised classification task trained on large labeled corpora.

Similarly, RAG hallucination detection has relied on natural language inference (NLI) benchmarks such as \textsc{TRUE}~\cite{honovich2022true} or fine-tuned evaluator LLMs such as \textsc{Lynx}~\cite{ravi2024lynx} and \textsc{Granite Guardian}~\cite{padhi2024granite} that check whether a generated answer is supported by a source document.
While accurate on their training domains, fine-tuned classifier LLMs require collecting new training data and retraining whenever the target policy or domain changes.

\subsubsection*{Hidden-State Probing and Training-Free $k$NN Guardrails}
To avoid the latency and retraining cost of a separate classifier LLM, recent work inspects the internal hidden states of the target LLM itself~\cite{zou2023representation}.
Classifiers trained on intermediate layer activations have been used to detect factual errors~\cite{azaria2023the,marks2023the}, steer models away from unsafe outputs at inference time (\textsc{SafeSwitch}~\cite{han2025safeswitch}), and detect indirect prompt injections~\cite{dong2026your}.

Non-parametric $k$NN search over hidden states~\cite{khandelwal2019generalization} avoids training even a linear probe, forming the basis of single-pass guardrails such as \textsc{SALLIE}~\cite{azov2026sallie} and \textsc{kNNGuard}~\cite{abdelfattah2026knnguard}.
However, Kale and Vishwanath~\cite{kale2026promptinjection} recently showed that hidden-state probes are sensitive to sequence-length shifts, and Dubanowska et al.~\cite{dubanowska2025representationbased} found that activation-based hallucination detectors degrade under distribution shifts, motivating long-form RAG detectors such as \textsc{FACTUM}~\cite{dassen2026factum} and our long-context audit of \textsc{kNNGuard}.

\subsection{\textsc{kNNGuard} Architecture}
Following Abdelfattah et al.~\cite{abdelfattah2026knnguard}, each task domain $d$ uses a labeled reference bank $\mathcal{B}^{(d)} = \{(x_i, y_i)\}_{i=1}^{2n}$ containing $n$ negative ($y_i=0$) and $n$ positive ($y_i=1$) examples, partitioned into index sets $C_0 = \{i : y_i = 0\}$ and $C_1 = \{i : y_i = 1\}$.
For the LLM activation branch, each raw input $x$ is prepended with a domain-specific system prompt to form $\tilde{x}$, instructing the LLM to classify negative (allowed) inputs as \texttt{on-topic} ($y=0$) and positive (rejected) inputs as \texttt{off-topic} ($y=1$).

\subsubsection*{Layer-Ensemble Activation Branch (\textsc{kNNGuard LE})}
Given an LLM with $L$ transformer decoder layers, \textsc{kNNGuard LE} (\emph{Layer Ensemble}) selects $M$ evenly spaced layers indexed by $\ell \in \{1, \dots, M\}$ along the depth of the network.
For each reference prompt $\tilde{x}_i$, let $\mathbf{h}_i^{(\ell)} \in \mathbb{R}^D$ denote its last-token hidden state at sampled layer $\ell$.
We first standardize $\mathbf{h}_i^{(\ell)}$ elementwise across the $D$ hidden dimensions using the bank mean $\boldsymbol{\mu}_{\text{bank}}^{(\ell)}$ and standard deviation $\boldsymbol{\sigma}_{\text{bank}}^{(\ell)}$ computed over all $2n$ bank examples at layer $\ell$:
\begin{equation}
\boldsymbol{\mu}_{\text{bank}}^{(\ell)} = \frac{1}{2n} \sum_{i=1}^{2n} \mathbf{h}_i^{(\ell)}, \quad \boldsymbol{\sigma}_{\text{bank}}^{(\ell)} = \sqrt{\frac{1}{2n} \sum_{i=1}^{2n} \left(\mathbf{h}_i^{(\ell)} - \boldsymbol{\mu}_{\text{bank}}^{(\ell)}\right)^2}, \quad \tilde{\mathbf{h}}_i^{(\ell)} = \frac{\mathbf{h}_i^{(\ell)} - \boldsymbol{\mu}_{\text{bank}}^{(\ell)}}{\boldsymbol{\sigma}_{\text{bank}}^{(\ell)}}.
\end{equation}
Each standardized bank vector is then divided by its L2 norm to obtain a unit vector $\hat{\mathbf{z}}_i^{(\ell)} = \tilde{\mathbf{h}}_i^{(\ell)} / \|\tilde{\mathbf{h}}_i^{(\ell)}\|_2$.

Let $\boldsymbol{\mu}_0^{(\ell)}$ and $\boldsymbol{\mu}_1^{(\ell)}$ denote the mean standardized activation vectors of the negative ($C_0$) and positive ($C_1$) bank subsets at layer $\ell$.
To measure how well layer $\ell$ separates the two classes, we compute its Fisher discriminant ratio $J_\ell$:
\begin{equation}
J_\ell = \frac{B_\ell}{W_\ell}, \quad B_\ell = \|\boldsymbol{\mu}_1^{(\ell)} - \boldsymbol{\mu}_0^{(\ell)}\|_2^2, \quad W_\ell = \sum_{c \in \{0,1\}} \frac{1}{|C_c|} \sum_{i \in C_c} \|\tilde{\mathbf{h}}_i^{(\ell)} - \boldsymbol{\mu}_c^{(\ell)}\|_2^2,
\end{equation}
and assign layer weights via a softmax across the $M$ sampled layers, $\alpha_\ell = \exp(J_\ell) / \sum_{m=1}^{M} \exp(J_m)$.

At inference time, for a new input $\tilde{x}$ with layer-$\ell$ last-token activation $\mathbf{q}^{(\ell)} \in \mathbb{R}^D$, we apply the same bank standardization $\tilde{\mathbf{q}}^{(\ell)} = (\mathbf{q}^{(\ell)} - \boldsymbol{\mu}_{\text{bank}}^{(\ell)}) / \boldsymbol{\sigma}_{\text{bank}}^{(\ell)}$ and divide by its L2 norm to obtain the unit query vector $\hat{\mathbf{q}}^{(\ell)} = \tilde{\mathbf{q}}^{(\ell)} / \|\tilde{\mathbf{q}}^{(\ell)}\|_2$.
We then compute the cosine distance $d_i^{(\ell)} = 1 - (\hat{\mathbf{z}}_i^{(\ell)})^\top \hat{\mathbf{q}}^{(\ell)}$ to each bank example $i \in \{1, \dots, 2n\}$. Let $\mathcal{N}_k^{(\ell)}(x) \subset \{1, \dots, 2n\}$ denote the indices of the $k$ bank examples with the smallest cosine distances $d_i^{(\ell)}$, with inverse-distance weights $w_i^{(\ell)} = 1 / d_i^{(\ell)}$:
\begin{equation}
s_\ell(x) = \frac{\sum_{i \in \mathcal{N}_k^{(\ell)}(x)} w_i^{(\ell)} y_i}{\sum_{i \in \mathcal{N}_k^{(\ell)}(x)} w_i^{(\ell)}}, \quad s_{\text{act}}(x) = \sum_{\ell=1}^{M} \alpha_\ell s_\ell(x).
\end{equation}
Intuitively, $s_{\text{act}}(x) \in [0, 1]$ aggregates evidence across the $M$ layers into a single risk score: within each layer $\ell$, the $k$ nearest bank neighbors vote on the query's class with weights inversely proportional to their distance from the query (so closer neighbors carry more weight), and the layer scores $s_\ell(x)$ are weighted by $\alpha_\ell$ so that layers with stronger class separation have greater influence.

\subsubsection*{Embedding Branch and Fused Ensemble (\textsc{kNNGuard FE})}
The embedding branch (\textsc{Embedding-kNN}) runs the same single-space $k$NN scoring using vectors from an external sentence encoder rather than LLM hidden states.
Specifically, we encode the raw reference prompts $\{x_i\}_{i=1}^{2n}$ (without the LLM system prompt) into L2-normalized vectors in $\mathbb{R}^{N}$ to form the embedding reference bank.
At inference time, a test query $x$ is encoded with the same sentence encoder and scored against its $k$ nearest cosine-distance neighbors in this embedding bank using inverse-distance weights (as in $s_\ell(x)$ in Equation~(3), without layer weighting since the encoder outputs a single vector per input) to obtain $s_{\text{emb}}(x)$.

The \emph{Fused Ensemble} (\textsc{kNNGuard FE}) combines $s_{\text{act}}(x)$ and $s_{\text{emb}}(x)$ using each branch's distance from the midpoint, $c_b(x) = |s_b(x) - 0.5|$ for $b \in \{\text{act}, \text{emb}\}$, and the confidence gap $\Delta(x) = |c_{\text{act}}(x) - c_{\text{emb}}(x)|$.
When one branch is more confident by at least a margin $\gamma$, Winner-Takes-All selects that branch; otherwise the two scores are blended proportionally:
\begin{equation}
s_{\text{FE}}(x) = \begin{cases}
s_{\arg\max_b c_b(x)}(x), & \text{if } \Delta(x) > \gamma, \\
\frac{c_{\text{act}}(x)s_{\text{act}}(x) + c_{\text{emb}}(x)s_{\text{emb}}(x)}{c_{\text{act}}(x) + c_{\text{emb}}(x)}, & \text{otherwise}.
\end{cases}
\end{equation}
Rather than choosing an arbitrary decision threshold to convert $s_{\text{act}}(x)$, $s_{\text{emb}}(x)$, and $s_{\text{FE}}(x)$ into binary predictions, we evaluate the continuous risk scores directly across all operating points using the Area Under the Receiver Operating Characteristic curve (AUROC).
Throughout our experiments, we follow the hyperparameter settings of Abdelfattah et al.~\cite{abdelfattah2026knnguard} ($n=50$ examples per class per bank, $M=9$ evenly spaced layers, $k=13$ nearest neighbors, and confidence margin $\gamma=0.1$), using \texttt{all-MiniLM-L6-v2} ($N=384$) when replicating the standard \textsc{kNNGuard} baseline.

\subsection{Target Models and Guardrail Benchmarks}
We evaluate two 4B open-weight models in native \texttt{BF16} precision without weight quantization: \texttt{Qwen/Qwen3.5-4B}~\cite{yang2025qwen3} ($L=32$ layers, hidden dimension $D=2560$) and \texttt{google/gemma-4-E4B-it}~\cite{team2025gemma} ($L=42$ layers, $D=2560$).
We evaluate two guardrail benchmarks:
\begin{itemize}
  \item \textbf{Domain 1 (Content Harm Moderation):} \texttt{nvidia/Aegis-AI-Content-Safety-Dataset-2.0}~\cite{ghosh2025aegis20}, which was also evaluated in \textsc{kNNGuard}~\cite{abdelfattah2026knnguard} and covers hate speech, violence, self-harm, harassment, and regulated advice.
  \item \textbf{Domain 2 (RAG Hallucination Detection):} \texttt{PatronusAI/HaluBench}~\cite{ravi2024lynx}, consisting of $(\texttt{passage}, \texttt{question}, \texttt{answer})$ triplets drawn from six QA and hallucination benchmarks (\texttt{HaluEval}~\cite{li2023halueval}, \texttt{DROP}~\cite{dua2019drop}, \texttt{PubMedQA}~\cite{jin2019pubmedqa}, \texttt{RAGTruth}~\cite{niu2023ragtruth}, \texttt{FinanceBench}~\cite{islam2023financebench}, and \texttt{COVID-QA}~\cite{moller2020covidqa}), where each triplet is labeled \texttt{PASS} ($y=0$, the answer is supported by the passage) or \texttt{FAIL} ($y=1$, the answer contains unsupported or contradicted claims).
\end{itemize}

For each benchmark, both the reference banks and the held-out evaluation buckets (\texttt{Short}: $\le 256$ tokens; \texttt{Medium}: $257$--$1000$ tokens; \texttt{Long}: $>1000$ tokens) are drawn from the benchmark's examples using strict disjoint partitioning:
\begin{itemize}
  \item In \textbf{\texttt{Aegis-2.0}}, which provides official \texttt{train}, \texttt{validation}, and \texttt{test} splits, we sample all reference banks strictly from the \texttt{train} split and build the evaluation buckets from the \texttt{test} and \texttt{validation} splits (plus held-out $>1000$-token \texttt{train} prompts not selected for any bank).
  To guard against near-duplicate prompts across splits, we also remove any evaluation prompt that matches a bank prompt exactly or shares its first $200$ characters.
  \item In \textbf{\texttt{HaluBench}}, which is released as a single split where multiple questions often share the same source passage, we partition the dataset by unique source passage: we sample the short reference bank ($\le 256$ tokens) from $100$ unique passages in \texttt{HaluEval} and the long reference bank ($>512$ tokens) from $100$ unique passages in \texttt{PubMedQA} and \texttt{RAGTruth}, and then exclude every example in \texttt{HaluBench} that shares a passage or question with either bank before forming the three evaluation buckets.
\end{itemize}

\section{Empirical Audit: Diagnosing Long-Context Degradation}
We first build the standard short-context reference banks $\mathcal{B}_{\text{short}}^{\text{safety}}$ and $\mathcal{B}_{\text{short}}^{\text{halu}}$ ($n=50$ per class, $\le 256$ tokens) and evaluate them on held-out \texttt{Short} ($\le 256$ tokens, $N=300$), \texttt{Medium} ($257$--$1000$ tokens, $N=260$--$300$), and \texttt{Long} ($>1000$ tokens, up to $6459$ tokens, $N=248$--$300$) test buckets.
Throughout our evaluation, we report the LLM activation branch (\textsc{kNNGuard LE}) alongside the Fused Ensemble (\textsc{kNNGuard FE}) for two reasons: (1) on content moderation (\texttt{Aegis-2.0}), inspecting \textsc{FE} alone masks a severe long-context collapse in the LLM activation branch because the $256$-token prefix embedding partially buffers the fused score, and (2) on RAG hallucination detection (\texttt{HaluBench}), we find that the embedding branch actively harms the fused ensemble once the reference bank is adapted, making \textsc{kNNGuard LE} alone stronger than \textsc{kNNGuard FE}.

\begin{table}[t]
\centering
\caption{Length-segmented AUROC of standard \textsc{kNNGuard LE} and \textsc{kNNGuard FE} (built from short $\le 256$-token banks $\mathcal{B}_{\text{short}}^{(d)}$) across native prompt-length regimes.}
\label{tab:step2_audit}
\resizebox{\textwidth}{!}{\begin{tabular}{lllcccc}
\toprule
Backbone & Benchmark & Branch & Short ($\le 256$) & Medium ($257$--$1000$) & Long ($>1000$) & $\Delta$ (Long$-$Short) \\
\midrule
Qwen3.5-4B & Aegis-2.0 (Safety) & kNNGuard LE & \textbf{0.777} & 0.736 & 0.587 & -0.190 \\
Qwen3.5-4B & Aegis-2.0 (Safety) & kNNGuard FE & \textbf{0.810} & 0.806 & 0.755 & -0.055 \\
Qwen3.5-4B & HaluBench (RAG) & kNNGuard LE & \textbf{0.995} & 0.725 & 0.598 & -0.397 \\
Qwen3.5-4B & HaluBench (RAG) & kNNGuard FE & \textbf{0.995} & 0.724 & 0.598 & -0.398 \\
Gemma-4-E4B-it & Aegis-2.0 (Safety) & kNNGuard LE & \textbf{0.706} & 0.628 & 0.521 & -0.186 \\
Gemma-4-E4B-it & Aegis-2.0 (Safety) & kNNGuard FE & \textbf{0.746} & 0.741 & 0.687 & -0.059 \\
Gemma-4-E4B-it & HaluBench (RAG) & kNNGuard LE & \textbf{0.992} & 0.695 & 0.513 & -0.479 \\
Gemma-4-E4B-it & HaluBench (RAG) & kNNGuard FE & \textbf{0.992} & 0.692 & 0.517 & -0.476 \\
\bottomrule
\end{tabular}
}
\end{table}

\begin{figure}[t]
\centering
\includegraphics[width=0.95\textwidth]{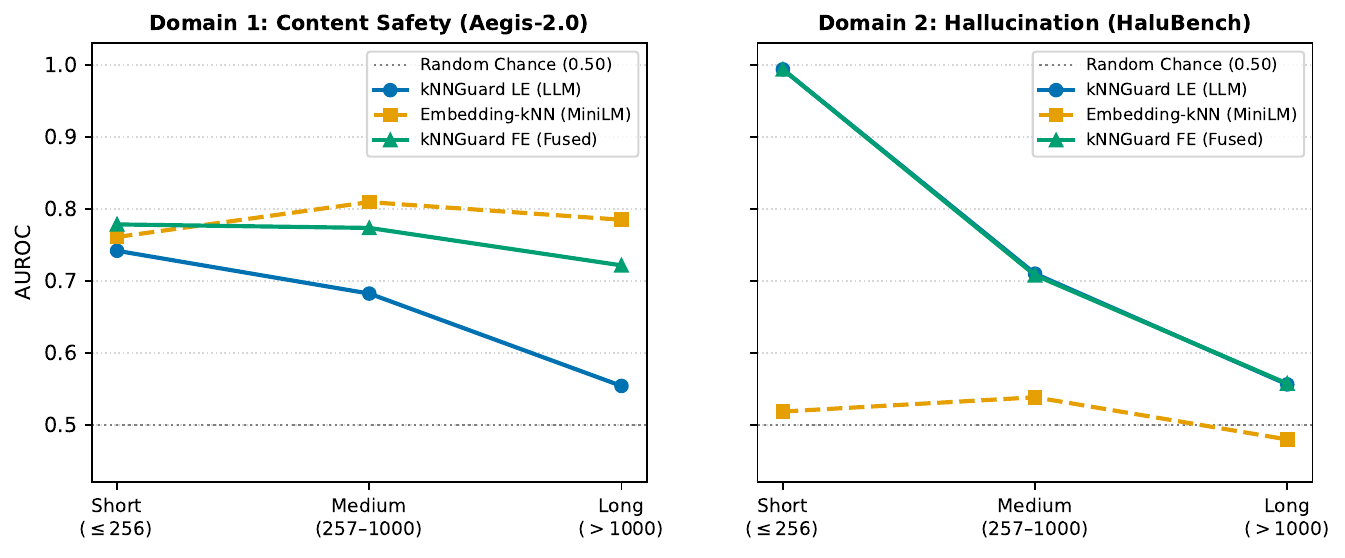}
\caption{Long-context AUROC degradation (averaged across \texttt{Qwen3.5-4B} and \texttt{Gemma-4-E4B-it}) across \texttt{Short} ($\le 256$), \texttt{Medium} ($257$--$1000$), and \texttt{Long} ($>1000$) test-query length buckets when using the standard short reference bank $\mathcal{B}_{\text{short}}^{(d)}$, for \textsc{kNNGuard LE} (LLM hidden states), \textsc{Embedding-kNN} (\texttt{MiniLM-256}), and fused \textsc{kNNGuard FE}.}
\label{fig:step2_degradation}
\end{figure}

\subsubsection*{Empirical Collapse Across Length Regimes}
As shown in Table~\ref{tab:step2_audit} and Figure~\ref{fig:step2_degradation}, standard \textsc{kNNGuard} achieves strong ranking accuracy when test inputs match the short length of the reference bank.
On \texttt{Short} ($\le 256$ tokens) inputs, \textsc{kNNGuard FE} reaches $0.810$ AUROC on \texttt{Qwen3.5-4B} and $0.746$ AUROC on \texttt{Gemma-4-E4B-it} for Content Safety, and $0.995$ AUROC and $0.992$ AUROC on \texttt{HaluBench}.

However, when the same short bank $\mathcal{B}_{\text{short}}^{(d)}$ is used to score \texttt{Medium} and \texttt{Long} inputs, the LLM activation branch (\textsc{kNNGuard LE}) degrades severely:
\begin{itemize}
  \item On \textbf{Content Safety (\texttt{Aegis-2.0})}, \textsc{kNNGuard LE} AUROC falls from $0.777 \to 0.736 \to 0.587$ on \texttt{Qwen3.5-4B} and from $0.706 \to 0.628 \to 0.521$ (near random chance) on \texttt{Gemma-4-E4B-it}.
  By contrast, the \texttt{MiniLM-256} embedding branch retains high AUROC on long safety prompts ($0.811$ and $0.758$), partially buffering the fused ensemble (\textsc{FE} AUROC drops only from $0.810 \to 0.755$ on \texttt{Qwen3.5-4B} and $0.746 \to 0.687$ on \texttt{Gemma-4-E4B-it}).

  Direct inspection of per-chunk embeddings on \texttt{Aegis-2.0 Long} explains why: harmful topic signal is present across the entire prompt (the first, middle, and final $256$-token chunks achieve $0.827$, $0.834$, and $0.808$ embedding AUROC on \texttt{Qwen3.5-4B}, and $0.853$, $0.875$, and $0.847$ on \texttt{Gemma-4-E4B-it}), and because \texttt{MiniLM-256} truncates every long test prompt to its first $256$ tokens, it compares a $256$-token query prefix against a $\le 256$-token bank without suffering from a sequence-length mismatch.
  \item On \textbf{RAG Hallucination Detection (\texttt{HaluBench})}, where \texttt{MiniLM-256} performs near random chance across all lengths ($0.460$--$0.558$ AUROC, including $0.513$--$0.523$ on \texttt{Short} inputs where no truncation occurs), the collapse of the LLM activation branch directly pulls down the fused ensemble: \textsc{kNNGuard FE} AUROC falls from $0.995 \to 0.724 \to 0.598$ on \texttt{Qwen3.5-4B} and from $0.992 \to 0.692 \to 0.517$ on \texttt{Gemma-4-E4B-it}.
\end{itemize}

\subsubsection*{Why Both Branches Fail on Long Inputs}
Inspecting the two branches separately explains the breakdown:
\begin{enumerate}
  \item \textbf{LLM Hidden-State Length Shift:} Standardizing long queries with the mean and standard deviation of $\mathcal{B}_{\text{short}}^{(d)}$ introduces a large representation offset as $T$ grows~\cite{kale2026promptinjection}.
  On \texttt{HaluBench}, the average layer-wise cosine similarity between the short bank's negative mean vector $\boldsymbol{\mu}_0^{\text{short}}$ and the mean vector of held-out negative queries in each length bucket drops from $0.840$ (\texttt{Short}) to $0.179$ (\texttt{Medium}) and $0.255$ (\texttt{Long}) on \texttt{Qwen3.5-4B}, and from $0.910$ (\texttt{Short}) to $0.063$ (\texttt{Medium}) and $0.248$ (\texttt{Long}) on \texttt{Gemma-4-E4B-it}.

  At the same time, the mean $k=13$ nearest-neighbor cosine distance increases from $0.706 \to 0.869$ (\texttt{Qwen3.5-4B}) and $0.676 \to 0.842$ (\texttt{Gemma-4-E4B-it}).
  \item \textbf{\texttt{MiniLM} Truncation Blindness:} Each \texttt{HaluBench} prompt is structured as:
  \begin{equation}
  \texttt{Passage: } \langle\text{retrieved context}\rangle \quad \texttt{Question: } \langle\text{query}\rangle \quad \texttt{Answer: } \langle\text{response}\rangle,
  \end{equation}
  placing the question and answer after the retrieved passage.
  The token index where \texttt{Question:} begins exceeds \texttt{MiniLM}'s $256$-token limit on $61.0\%$ (\texttt{Qwen3.5-4B}) to $66.0\%$ (\texttt{Gemma-4-E4B-it}) of \texttt{Medium} inputs and $\mathbf{100.0\%}$ of \texttt{Long} inputs, cutting off the question and answer before encoding (\texttt{MiniLM-256} Long AUROC is $0.460$--$0.499$).
\end{enumerate}

\section{Training-Free Long-Context Adaptations}
To address these failure mechanisms without model fine-tuning, we evaluate three modular, training-free adaptations:

\subsection{Length-Routed Dual Reference Banks}
Because last-token hidden states shift systematically as sequence length increases, forcing a single short reference bank $\mathcal{B}_{\text{short}}^{(d)}$ ($\le 256$ tokens) to score long queries distorts both the per-layer standardization step and nearest-neighbor retrieval.
To include long-context examples in the reference bank without collecting more than $n=50$ examples per class per bank, we compare two strategies:
\begin{itemize}
  \item \textbf{Mixed-Length Bank ($\mathcal{B}_{\text{mixed}}^{(d)}$):} A single $100$-example reference bank combining $25$ short ($\le 256$ tokens) and $25$ long ($>512$ tokens) examples per class.\\
  While simple, pooling short and long prompts into a single bank computes a single standardization mean $\boldsymbol{\mu}_{\text{bank}}^{(\ell)}$ and standard deviation $\boldsymbol{\sigma}_{\text{bank}}^{(\ell)}$ across two shifted activation clusters, blurring class separation on short inputs.
  \item \textbf{Length-Routed Dual Bank ($\mathcal{B}_{\text{short}}^{(d)}, \mathcal{B}_{\text{long}}^{(d)}$):} Alongside the short bank $\mathcal{B}_{\text{short}}^{(d)}$ ($n=50$ per class, $\le 256$ tokens), we pre-compute a separate \textbf{Long-Context Reference Bank} $\mathcal{B}_{\text{long}}^{(d)}$ ($n=50$ per class sampled from $>512$-token training prompts in \texttt{Aegis-2.0} and $>512$-token unique passages in \texttt{HaluBench}).\\
  Each bank maintains its own standardization parameters $(\boldsymbol{\mu}_{\text{bank}}^{(\ell)}, \boldsymbol{\sigma}_{\text{bank}}^{(\ell)})$ and layer weights $\alpha_\ell^{\text{short}}$ vs.\ $\alpha_\ell^{\text{long}}$.
  At inference time, as soon as the query is tokenized ($T = |\texttt{input\_ids}|$), we route the query to the length-matched bank:
  \begin{equation}
  \mathcal{B}_{\text{active}}^{(d)}(x) = \begin{cases}
  \mathcal{B}_{\text{short}}^{(d)}, & \text{if } T \le 256, \\
  \mathcal{B}_{\text{long}}^{(d)}, & \text{if } T > 256.
  \end{cases}
  \end{equation}
\end{itemize}
This routing adds no latency overhead, requires no extra forward passes, keeps the short and long activation distributions separate rather than averaging their statistics in a single bank, and preserves $100\%$ of short-prompt performance.

\subsection{Long-Context Embedding Replacement}
To eliminate \texttt{MiniLM}'s $256$-token truncation blindness while maintaining a lightweight embedding branch, we replace \texttt{all-MiniLM-L6-v2} with \texttt{Qwen/Qwen3-Embedding-0.6B}~\cite{zhang2025qwen3embedding}, which produces $1024$-dimensional embeddings ($N=1024$) and supports up to $32768$ tokens natively.
Because every prompt in our benchmarks fits well within this context window (up to $6459$ tokens), the embedding branch encodes the full prompt without truncation.

\subsection{Strided Window Pooling}
\label{sec:window_pooling}
Even when the reference bank is length-matched ($\mathcal{B}_{\text{long}}^{(d)}$) and both branches process all $T$ tokens, compressing a $2000$-to-$6000$ token prompt into a single end-of-sequence vector dilutes localized unsafe statements or single-sentence factual contradictions buried inside a long benign context.

To recover localized signal without running multiple LLM forward passes, we score long prompts at a $256$-token granularity according to each branch's attention mechanism:
\begin{itemize}
  \item \textbf{LLM Activation Branch (\textsc{kNNGuard LE}):} Because decoder-only LLMs use causal (left-to-right) self-attention, a single prefill pass over the full $T$-token prompt already computes intermediate representations at every token position.\\
  We read out the hidden states at every $256$-th token ($t \in \{256, 512, \dots, T\}$, i.e., the last token of each cumulative $256$-token prefix) across the $M=9$ selected layers and score each position against $\mathcal{B}_{\text{active}}^{(d)}(x)$.
  \item \textbf{Embedding Branch (\textsc{Embedding-kNN}):} Because the sentence encoder (\texttt{Qwen3-Embedding-0.6B}) outputs a single vector per input text, we split the prompt into sliding windows of roughly $256$ tokens with a $64$-token overlap (so that statements straddling a boundary are not cut in half) and score each chunk against the embedding bank.\\
  On \texttt{HaluBench}, we apply this sliding window to the retrieved \texttt{Passage} and append the \texttt{Question} and \texttt{Answer} to every passage chunk (which easily fits within \texttt{Qwen3-Embedding-0.6B}'s context window) so that each chunk embedding includes both the passage segment and the answer being verified.
\end{itemize}
For both branches, we aggregate the window-level risk scores into a single branch score ($s_{\text{act}}(x)$ and $s_{\text{emb}}(x)$) using \texttt{Top-2 Mean} pooling (averaging the two highest-scoring windows, or taking the single window score when $W=1$).
Averaging the top two windows captures localized high-risk segments while being less sensitive to single-window outliers than taking the raw maximum across windows; the resulting branch scores $s_{\text{act}}(x)$ and $s_{\text{emb}}(x)$ are then combined via Equation~(4).

\section{Experimental Results and Ablation Study}
\label{sec:results}
\begin{figure}[t]
\centering
\includegraphics[width=0.95\textwidth]{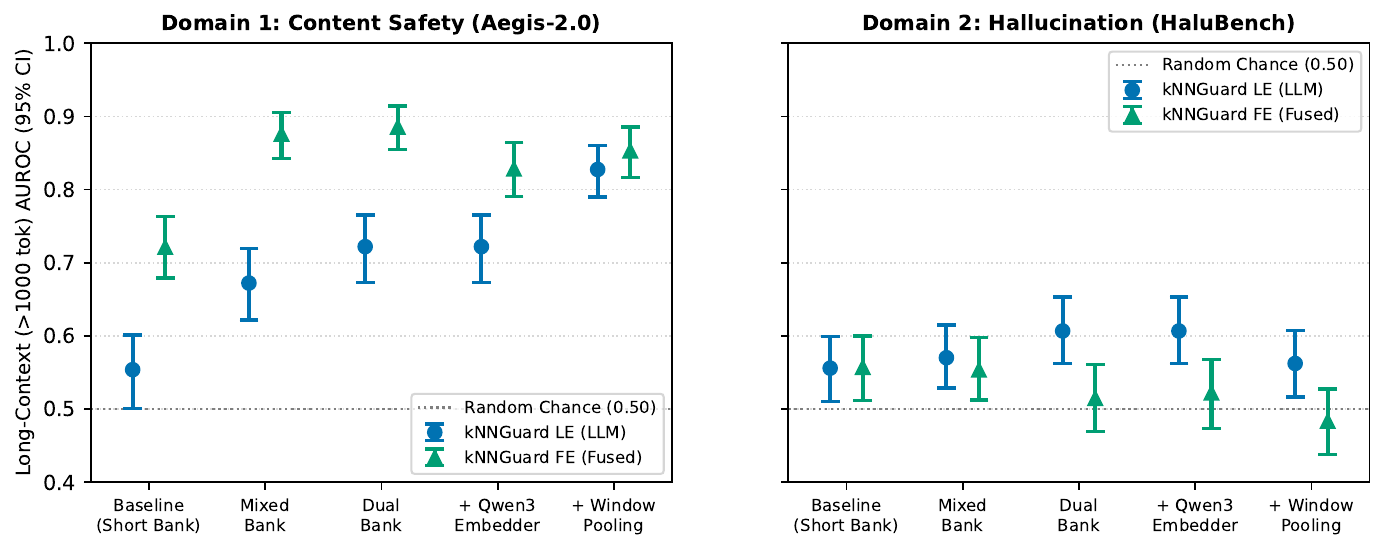}
\caption{Long-context ($>1000$ tokens) AUROC and $95\%$ bootstrap confidence intervals (averaged across \texttt{Qwen3.5-4B} and \texttt{Gemma-4-E4B-it}) of \textsc{kNNGuard LE} (LLM hidden states) and \textsc{kNNGuard FE} (fused ensemble) across training-free adaptations on Content Safety (\texttt{Aegis-2.0}) and RAG Hallucination (\texttt{HaluBench}).}
\label{fig:ablation_ladder}
\end{figure}

Table~\ref{tab:step3_fixes} (with Figure~\ref{fig:ablation_ladder}) and Table~\ref{tab:step3_medium} report our progressive ablation study on the \texttt{Long} ($>1000$-token) and \texttt{Medium} ($257$--$1000$-token) evaluation buckets across both domains, reporting \textsc{kNNGuard LE}, \textsc{Embedding-kNN}, and \textsc{kNNGuard FE} AUROC with $95\%$ bootstrap confidence intervals.

\begin{table}[t]
\centering
\caption{Progressive ablation study on \texttt{Long} ($>1000$-token) inputs averaged across \texttt{Qwen3.5-4B} and \texttt{Gemma-4-E4B-it}.}
\label{tab:step3_fixes}
\resizebox{\textwidth}{!}{\begin{tabular}{llcccc}
\toprule
Benchmark & Configuration & LE AUROC (95\% CI) & Emb AUROC & FE AUROC (95\% CI) & $\Delta$LE / $\Delta$FE \\
\midrule
Aegis-2.0 (Safety) & Baseline (Short Bank) & 0.554 [0.501, 0.601] & 0.785 & 0.721 [0.679, 0.763] & +0.000 / +0.000 \\
Aegis-2.0 (Safety) & Mixed-Length Bank & 0.672 [0.622, 0.720] & \textbf{0.895} & 0.876 [0.843, 0.906] & +0.118 / +0.155 \\
Aegis-2.0 (Safety) & Length-Routed Dual Bank & 0.722 [0.673, 0.765] & 0.894 & \textbf{0.886} [0.855, 0.914] & +0.168 / +0.164 \\
Aegis-2.0 (Safety) & + Qwen3-Embedding-0.6B & 0.722 [0.673, 0.765] & 0.847 & 0.828 [0.790, 0.864] & +0.168 / +0.107 \\
Aegis-2.0 (Safety) & + Strided Window Pooling & \textbf{0.828} [0.790, 0.861] & 0.853 & 0.853 [0.817, 0.885] & +0.274 / +0.132 \\
\midrule
HaluBench (RAG) & Baseline (Short Bank) & 0.556 [0.510, 0.600] & 0.479 & \textbf{0.557} [0.512, 0.600] & +0.000 / +0.000 \\
HaluBench (RAG) & Mixed-Length Bank & 0.570 [0.529, 0.615] & 0.476 & 0.554 [0.512, 0.598] & +0.014 / -0.004 \\
HaluBench (RAG) & Length-Routed Dual Bank & \textbf{0.607} [0.563, 0.653] & \textbf{0.489} & 0.515 [0.470, 0.561] & +0.051 / -0.042 \\
HaluBench (RAG) & + Qwen3-Embedding-0.6B & \textbf{0.607} [0.563, 0.653] & 0.477 & 0.522 [0.474, 0.568] & +0.051 / -0.035 \\
HaluBench (RAG) & + Strided Window Pooling & 0.562 [0.516, 0.607] & 0.467 & 0.484 [0.438, 0.528] & +0.006 / -0.074 \\
\bottomrule
\end{tabular}
}
\end{table}

\begin{table}[t]
\centering
\caption{Progressive ablation study on \texttt{Medium} ($257$--$1000$-token) inputs averaged across \texttt{Qwen3.5-4B} and \texttt{Gemma-4-E4B-it}.}
\label{tab:step3_medium}
\resizebox{\textwidth}{!}{\begin{tabular}{llcccc}
\toprule
Benchmark & Configuration & LE AUROC (95\% CI) & Emb AUROC & FE AUROC (95\% CI) & $\Delta$LE / $\Delta$FE \\
\midrule
Aegis-2.0 (Safety) & Baseline (Short Bank) & 0.682 [0.634, 0.732] & 0.809 & 0.773 [0.733, 0.812] & +0.000 / +0.000 \\
Aegis-2.0 (Safety) & Mixed-Length Bank & 0.716 [0.675, 0.760] & 0.871 & 0.862 [0.830, 0.896] & +0.034 / +0.088 \\
Aegis-2.0 (Safety) & Length-Routed Dual Bank & 0.782 [0.742, 0.820] & \textbf{0.874} & \textbf{0.870} [0.836, 0.901] & +0.099 / +0.096 \\
Aegis-2.0 (Safety) & + Qwen3-Embedding-0.6B & 0.782 [0.742, 0.820] & 0.818 & 0.829 [0.795, 0.863] & +0.099 / +0.056 \\
Aegis-2.0 (Safety) & + Strided Window Pooling & \textbf{0.808} [0.771, 0.850] & 0.823 & 0.823 [0.788, 0.858] & +0.126 / +0.050 \\
\midrule
HaluBench (RAG) & Baseline (Short Bank) & 0.710 [0.669, 0.749] & 0.538 & 0.708 [0.667, 0.747] & +0.000 / +0.000 \\
HaluBench (RAG) & Mixed-Length Bank & \textbf{0.853} [0.823, 0.880] & 0.499 & \textbf{0.837} [0.806, 0.868] & +0.143 / +0.130 \\
HaluBench (RAG) & Length-Routed Dual Bank & 0.649 [0.605, 0.693] & 0.533 & 0.610 [0.567, 0.654] & -0.060 / -0.098 \\
HaluBench (RAG) & + Qwen3-Embedding-0.6B & 0.649 [0.605, 0.693] & \textbf{0.552} & 0.599 [0.553, 0.644] & -0.060 / -0.109 \\
HaluBench (RAG) & + Strided Window Pooling & 0.632 [0.587, 0.676] & 0.548 & 0.568 [0.520, 0.613] & -0.078 / -0.140 \\
\bottomrule
\end{tabular}
}
\end{table}

\subsubsection*{Finding 1: Length-Matched Banks and Window Pooling Restore Long-Context Safety Ranking}
On \textbf{Content Safety (\texttt{Aegis-2.0})}, each adaptation progressively improves ranking accuracy across both the \texttt{Long} (Table~\ref{tab:step3_fixes}, Figure~\ref{fig:ablation_ladder}) and \texttt{Medium} (Table~\ref{tab:step3_medium}) buckets without model fine-tuning:
\begin{itemize}
  \item \textbf{Reference-Bank Adaptation:} Replacing the short bank with a \textbf{Mixed-Length Bank} substantially improves both \textsc{kNNGuard LE} and \textsc{kNNGuard FE} on long and medium inputs, but slightly degrades short-prompt AUROC ($0.810 \to 0.791$ on \texttt{Qwen3.5-4B} and $0.746 \to 0.700$ on \texttt{Gemma-4-E4B-it}) because short and long activations are pooled into a single standardization distribution.\\
  The \textbf{Length-Routed Dual Bank} avoids this tradeoff, achieving the highest fused \textsc{kNNGuard FE} AUROC on both \texttt{Long} (Table~\ref{tab:step3_fixes}) and \texttt{Medium} (Table~\ref{tab:step3_medium}) safety prompts while preserving $100\%$ of short-prompt performance.
  \item \textbf{Full-Sequence vs.\ Windowed Pooling:} In the \texttt{+ Qwen3-Embedding-0.6B} configuration of Table~\ref{tab:step3_fixes}, encoding an entire $>1000$-token safety prompt into a single vector slightly reduces \textsc{Embedding-kNN} AUROC compared to \texttt{MiniLM-256}'s $256$-token prefix, because \texttt{MiniLM-256} truncates both the reference bank and test queries to the exact same $256$-token length whereas full-sequence \texttt{Qwen3-Embedding-0.6B} compares $>512$-token bank prompts against $1000$--$6459$ token test prompts compressed into one vector.\\
  Adding \textbf{Strided Window Pooling} (\texttt{+ Strided Window Pooling} in Tables~\ref{tab:step3_fixes} and~\ref{tab:step3_medium}) resolves single-vector dilution for both branches and delivers the largest gain for the LLM activation branch (\textsc{kNNGuard LE} reaches its peak AUROC on both \texttt{Long} and \texttt{Medium} safety prompts), with \texttt{Top-2 Mean} outperforming raw \texttt{Max} pooling on \texttt{Long} inputs ($0.828$ vs.\ $0.815$ \textsc{LE} AUROC and $0.853$ vs.\ $0.834$ \textsc{FE} AUROC).
\end{itemize}

\subsubsection*{Finding 2: Why Hallucination Detection Diverges Between Embeddings and LLM Hidden States}
Comparing \texttt{HaluBench} against \texttt{Aegis-2.0} in Tables~\ref{tab:step3_fixes} and~\ref{tab:step3_medium} reveals a sharp contrast between the sentence-embedding branch and the LLM hidden-state branch:
\begin{itemize}
  \item \textbf{Sentence-Embedding $k$NN Fails Across All Context Lengths on Hallucination Detection:} While \textsc{Embedding-kNN} separates safe and unsafe prompts effectively on \texttt{Aegis-2.0}, it remains near random chance ($\approx 0.47$--$0.55$ AUROC) across \emph{every} length bucket and configuration on \texttt{HaluBench} (Tables~\ref{tab:step3_fixes} and~\ref{tab:step3_medium}), even when $256$-token truncation is completely removed via full-sequence \texttt{Qwen3-Embedding-0.6B} or QA-anchored $256$-token chunking (and even on \texttt{Short} $\le 256$-token prompts where no truncation occurs).\\
  Detecting a hallucination depends on verifying logical consistency across tokens between the \texttt{Answer} and the \texttt{Passage}, which is too fine-grained to be captured by $k$NN similarity over static sentence embeddings.\\
  As a result, fusing the uninformative embedding branch with \textsc{kNNGuard LE} actively degrades \textsc{kNNGuard FE} on \texttt{HaluBench} whenever \textsc{LE} is above chance.
  \item \textbf{LLM Hidden States (\textsc{kNNGuard LE}) Succeed on Short and Medium Passages but Dilute on $>1000$-Token Passages:} Unlike the external sentence encoder, the LLM processes the verification system prompt via self-attention across the \texttt{Passage}, \texttt{Question}, and \texttt{Answer}.\\
  Consequently, $k$NN over the LLM's final-token hidden states (\textsc{kNNGuard LE}) achieves near-perfect separation on \texttt{Short} inputs ($\le 256$ tokens, Table~\ref{tab:step2_audit}) and recovers strong accuracy on \texttt{Medium} inputs ($257$--$1000$ tokens) once longer examples are included in the Mixed-Length Bank ($\text{LE AUROC} = 0.853$, Table~\ref{tab:step3_medium}; the Mixed-Length Bank outperforms the Dual Bank here because medium inputs span the $257$--$512$ token transition between the short and long banks).\\
  On \texttt{Long} ($>1000$-token) passages (Table~\ref{tab:step3_fixes}), however, the \texttt{Length-Routed Dual Bank} yields only a modest improvement for \textsc{kNNGuard LE} ($0.556 \to 0.607$), and \textbf{Strided Window Pooling} reduces \textsc{LE} AUROC ($0.562$): because the \texttt{Answer} being verified sits at the very end of the prompt (Equation~(5)), intermediate LLM hidden states read out at $t = 256, 512, \dots$ precede the \texttt{Answer} and cannot evaluate its consistency, while at the final token $t = T$, compressing a multi-thousand-token passage into a single hidden-state vector leaves too weak a signal for $k$NN over $100$ reference vectors.
\end{itemize}

\section{Discussion, Limitations, and Conclusion}
\subsubsection*{Practical Deployment Recommendations}
Our findings yield three practical guidelines for deploying training-free activation guardrails:
\begin{enumerate}
  \item \textbf{Deploy a Length-Routed Dual Bank by default:} Maintaining a separate long-context reference bank alongside the short bank is simple to deploy, adds no runtime overhead, and substantially improves medium- and long-context content moderation (Tables~\ref{tab:step3_fixes} and~\ref{tab:step3_medium}) with no downsides on short prompts.
  \item \textbf{Use Strided Window Pooling for long-context content moderation:} Sampling intermediate hidden states every $256$ tokens during the single prefill pass provides the strongest long-context activation performance (\textsc{kNNGuard LE}) on \texttt{Aegis-2.0} (Table~\ref{tab:step3_fixes}) without requiring additional LLM forward passes.
  \item \textbf{Disable Embedding-$k$NN for RAG hallucination detection:} Because $k$NN over static sentence embeddings cannot verify logical consistency between an answer and a source passage, RAG hallucination guardrails should rely on the LLM activation branch (\textsc{kNNGuard LE}) with length-matched banks rather than embedding fusion.
\end{enumerate}

\subsubsection*{Limitations and Conclusion}
Our evaluation focuses on two 4B open-weight model families (\texttt{Qwen3.5-4B} and \texttt{Gemma-4-E4B-it}) and two guardrail benchmarks up to $6.5$K tokens.

While length-routed dual banks and strided window pooling restore long-context content moderation on \texttt{Aegis-2.0} ($0.828$ \textsc{LE} AUROC, $0.886$ \textsc{FE} AUROC) and medium-context hallucination detection on \texttt{HaluBench} ($0.824$--$0.882$ \textsc{LE} AUROC), extending training-free $k$NN guardrails to multi-thousand-token financial and medical RAG verification will require representations that directly capture passage-to-answer consistency beyond static reference banks.

\bibliographystyle{plain}
\bibliography{references}

@misc{abdelfattah2026knnguard,
  title = {kNNGuard: Turning LLM Hidden Activations into a Training-Free Configurable Guardrail},
  author = {Mahmoud Abdelfattah and Hamid Nasiri and Peter Garraghan},
  year = {2026},
  eprint = {2607.02072},
  archivePrefix = {arXiv},
  primaryClass = {cs.LG},
  url = {https://arxiv.org/abs/2607.02072}
}

@misc{kale2026promptinjection,
  title = {Prompt-Injection Probes Learn the Conversation, Not Just the Injection},
  author = {Kaan Kale and Sriram Vishwanath},
  year = {2026},
  eprint = {2610.18408},
  archivePrefix = {arXiv},
  primaryClass = {cs.CL},
  url = {https://arxiv.org/abs/2610.18408}
}

@misc{azov2026sallie,
  title = {SALLIE: Generation-Free Hidden-State Detection of Jailbreaks and Prompt Injections Across Text and Vision},
  author = {Guy Azov and Ofer Rivlin and Guy Shtar},
  year = {2026},
  eprint = {2604.06247},
  archivePrefix = {arXiv},
  primaryClass = {cs.CR},
  url = {https://arxiv.org/abs/2604.06247}
}

@misc{ravi2024lynx,
  title = {Lynx: An Open Source Hallucination Evaluation Model},
  author = {Selvan Sunitha Ravi and Bartosz Mielczarek and Anand Kannappan and Douwe Kiela and Rebecca Qian},
  year = {2024},
  eprint = {2407.08488},
  archivePrefix = {arXiv},
  primaryClass = {cs.AI},
  url = {https://arxiv.org/abs/2407.08488}
}

@misc{ghosh2025aegis20,
  title = {Aegis2.0: A Diverse AI Safety Dataset and Risks Taxonomy for Alignment of LLM Guardrails},
  author = {Shaona Ghosh and Prasoon Varshney and Makesh Narsimhan Sreedhar and Aishwarya Padmakumar and Traian Rebedea and Jibin Rajan Varghese and Christopher Parisien},
  year = {2025},
  eprint = {2501.09004},
  archivePrefix = {arXiv},
  primaryClass = {cs.CL},
  url = {https://arxiv.org/abs/2501.09004}
}

@inproceedings{reimers2019sentencebert,
  title = {Sentence-BERT: Sentence Embeddings using Siamese BERT-Networks},
  author = {Nils Reimers and Iryna Gurevych},
  booktitle = {Proceedings of the 2019 Conference on Empirical Methods in Natural Language Processing (EMNLP)},
  year = {2019},
  url = {https://arxiv.org/abs/1908.10084}
}

@misc{zhang2025qwen3embedding,
  title = {Qwen3 Embedding: Advancing Text Embedding and Reranking Through Foundation Models},
  author = {Yanzhao Zhang and Mingxin Li and Dingkun Long and Xin Zhang and Huan Lin and Baosong Yang and Pengjun Xie and An Yang and Dayiheng Liu and Junyang Lin and Fei Huang and Jingren Zhou},
  year = {2025},
  eprint = {2506.05176},
  archivePrefix = {arXiv},
  primaryClass = {cs.CL},
  url = {https://arxiv.org/abs/2506.05176}
}

@misc{metallama2025llamaguard4,
  title = {Llama Guard 4 (12B): Native Multimodal Safety Classifier},
  author = {{Meta AI}},
  year = {2025},
  howpublished = {\url{https://huggingface.co/meta-llama/Llama-Guard-4-12B}},
  url = {https://huggingface.co/meta-llama/Llama-Guard-4-12B}
}

@misc{bai2023longbench,
  title = {LongBench: A Bilingual, Multitask Benchmark for Long Context Understanding},
  author = {Yushi Bai and Xin Lv and Jiajie Zhang and Hongchang Lyu and Jiankai Tang and Zhidian Huang and Zhengxiao Du and Xiao Liu and Aohan Zeng and Lei Hou and Yuxiao Dong and Jie Tang and Juanzi Li},
  year = {2023},
  eprint = {2308.14508},
  archivePrefix = {arXiv},
  primaryClass = {cs.CL},
  url = {https://arxiv.org/abs/2308.14508}
}

@misc{rebedea2023nemo,
  title = {NeMo Guardrails: A Toolkit for Controllable and Safe LLM Applications with Programmable Rails},
  author = {Traian Rebedea and Razvan Dinu and Makesh Sreedhar and Christopher Parisien and Jonathan Cohen},
  year = {2023},
  eprint = {2310.10501},
  archivePrefix = {arXiv},
  primaryClass = {cs.CL},
  url = {https://arxiv.org/abs/2310.10501}
}

@misc{zou2023representation,
  title = {Representation Engineering: A Top-Down Approach to AI Transparency},
  author = {Andy Zou and Long Phan and Sarah Chen and James Campbell and Phillip Guo and Richard Ren and Alexander Pan and Xuwang Yin and Mantas Mazeika and Ann-Kathrin Dombrowski and Shashwat Goel and Nathaniel Li and Michael J. Byun and Zifan Wang and Alex Mallen and Steven Basart and Sanmi Koyejo and Dawn Song and Matt Fredrikson and J. Zico Kolter and Dan Hendrycks},
  year = {2023},
  eprint = {2310.01405},
  archivePrefix = {arXiv},
  primaryClass = {cs.LG},
  url = {https://arxiv.org/abs/2310.01405}
}

@misc{marks2023the,
  title = {The Geometry of Truth: Emergent Linear Structure in Large Language Model Representations of True/False Datasets},
  author = {Samuel Marks and Max Tegmark},
  year = {2023},
  eprint = {2310.06824},
  archivePrefix = {arXiv},
  primaryClass = {cs.AI},
  url = {https://arxiv.org/abs/2310.06824}
}

@misc{azaria2023the,
  title = {The Internal State of an LLM Knows When It's Lying},
  author = {Amos Azaria and Tom Mitchell},
  year = {2023},
  eprint = {2304.13734},
  archivePrefix = {arXiv},
  primaryClass = {cs.CL},
  url = {https://arxiv.org/abs/2304.13734}
}

@misc{khandelwal2019generalization,
  title = {Generalization through Memorization: Nearest Neighbor Language Models},
  author = {Urvashi Khandelwal and Omer Levy and Dan Jurafsky and Luke Zettlemoyer and Mike Lewis},
  year = {2019},
  eprint = {1911.00172},
  archivePrefix = {arXiv},
  primaryClass = {cs.CL},
  url = {https://arxiv.org/abs/1911.00172}
}

@misc{su2021roformer,
  title = {RoFormer: Enhanced Transformer with Rotary Position Embedding},
  author = {Jianlin Su and Yu Lu and Shengfeng Pan and Ahmed Murtadha and Bo Wen and Yunfeng Liu},
  year = {2021},
  eprint = {2104.09864},
  archivePrefix = {arXiv},
  primaryClass = {cs.CL},
  url = {https://arxiv.org/abs/2104.09864}
}

@misc{liu2023lost,
  title = {Lost in the Middle: How Language Models Use Long Contexts},
  author = {Nelson F. Liu and Kevin Lin and John Hewitt and Ashwin Paranjape and Michele Bevilacqua and Fabio Petroni and Percy Liang},
  year = {2023},
  eprint = {2307.03172},
  archivePrefix = {arXiv},
  primaryClass = {cs.CL},
  url = {https://arxiv.org/abs/2307.03172}
}

@misc{niu2023ragtruth,
  title = {RAGTruth: A Hallucination Corpus for Developing Trustworthy Retrieval-Augmented Language Models},
  author = {Cheng Niu and Yuanhao Wu and Juno Zhu and Siliang Xu and Kashun Shum and Randy Zhong and Juntong Song and Tong Zhang},
  year = {2023},
  eprint = {2401.00396},
  archivePrefix = {arXiv},
  primaryClass = {cs.CL},
  url = {https://arxiv.org/abs/2401.00396}
}

@misc{islam2023financebench,
  title = {FinanceBench: A New Benchmark for Financial Question Answering},
  author = {Pranab Islam and Anand Kannappan and Douwe Kiela and Rebecca Qian and Nino Scherrer and Bertie Vidgen},
  year = {2023},
  eprint = {2311.11944},
  archivePrefix = {arXiv},
  primaryClass = {cs.CL},
  url = {https://arxiv.org/abs/2311.11944}
}

@misc{li2023halueval,
  title = {HaluEval: A Large-Scale Hallucination Evaluation Benchmark for Large Language Models},
  author = {Junyi Li and Xiaoxue Cheng and Wayne Xin Zhao and Jian-Yun Nie and Ji-Rong Wen},
  year = {2023},
  eprint = {2305.11747},
  archivePrefix = {arXiv},
  primaryClass = {cs.CL},
  url = {https://arxiv.org/abs/2305.11747}
}

@misc{jin2019pubmedqa,
  title = {PubMedQA: A Dataset for Biomedical Research Question Answering},
  author = {Qiao Jin and Bhuwan Dhingra and Zhengping Liu and William W. Cohen and Xinghua Lu},
  year = {2019},
  eprint = {1909.06146},
  archivePrefix = {arXiv},
  primaryClass = {cs.CL},
  url = {https://arxiv.org/abs/1909.06146}
}

@misc{dua2019drop,
  title = {DROP: A Reading Comprehension Benchmark Requiring Discrete Reasoning Over Paragraphs},
  author = {Dheeru Dua and Yizhong Wang and Pradeep Dasigi and Gabriel Stanovsky and Sameer Singh and Matt Gardner},
  year = {2019},
  eprint = {1903.00161},
  archivePrefix = {arXiv},
  primaryClass = {cs.CL},
  url = {https://arxiv.org/abs/1903.00161}
}

@inproceedings{moller2020covidqa,
  title = {COVID-QA: A Question Answering Dataset for COVID-19},
  author = {Timo M{\"o}ller and Anthony Reina and Raghavan Jayakumar and Malte Pietsch},
  booktitle = {Proceedings of the 1st Workshop on NLP for COVID-19 at ACL 2020},
  year = {2020},
  url = {https://aclanthology.org/2020.nlpcovid19-acl.18/}
}

@misc{wang2020minilm,
  title = {MiniLM: Deep Self-Attention Distillation for Task-Agnostic Compression of Pre-Trained Transformers},
  author = {Wenhui Wang and Furu Wei and Li Dong and Hangbo Bao and Nan Yang and Ming Zhou},
  year = {2020},
  eprint = {2002.10957},
  archivePrefix = {arXiv},
  primaryClass = {cs.CL},
  url = {https://arxiv.org/abs/2002.10957}
}

@misc{yang2025qwen3,
  title = {Qwen3 Technical Report},
  author = {An Yang and Anfeng Li and Baosong Yang and Beichen Zhang and Binyuan Hui and Bo Zheng and Bowen Yu and Chang Gao and Chengen Huang and Chenxu Lv and Chujie Zheng and Dayiheng Liu and Fan Zhou and Fei Huang and Jingren Zhou and Junyang Lin and others},
  year = {2025},
  eprint = {2505.09388},
  archivePrefix = {arXiv},
  primaryClass = {cs.CL},
  url = {https://arxiv.org/abs/2505.09388}
}

@misc{team2025gemma,
  title = {Gemma 3 Technical Report},
  author = {Gemma Team and Aishwarya Kamath and Johan Ferret and Shreya Pathak and Nino Vieillard and Ramona Merhej and Sarah Perrin and others},
  year = {2025},
  eprint = {2503.19786},
  archivePrefix = {arXiv},
  primaryClass = {cs.CL},
  url = {https://arxiv.org/abs/2503.19786}
}

@misc{dong2026your,
  title = {Your Agentic LLMs Secretly Encode Indirect Prompt-Injection Exposure in Hidden States},
  author = {Jianshuo Dong and Yiming Liu and Maosen Zhang and Nan Deng and Peng Xu and Xiaoping Zhang and Tianwei Zhang and Jie Zhang and Han Qiu},
  year = {2026},
  eprint = {2608.02657},
  archivePrefix = {arXiv},
  primaryClass = {cs.CR},
  url = {https://arxiv.org/abs/2608.02657}
}

@misc{lewis2020retrievalaugmented,
  title = {Retrieval-Augmented Generation for Knowledge-Intensive NLP Tasks},
  author = {Patrick Lewis and Ethan Perez and Aleksandra Piktus and Fabio Petroni and Vladimir Karpukhin and Naman Goyal and Heinrich K{\"u}ttler and Mike Lewis and Wen-tau Yih and Tim Rockt{\"a}schel and Sebastian Riedel and Douwe Kiela},
  year = {2020},
  eprint = {2005.11401},
  archivePrefix = {arXiv},
  primaryClass = {cs.CL},
  url = {https://arxiv.org/abs/2005.11401}
}

@misc{honovich2022true,
  title = {TRUE: Re-evaluating Factual Consistency Evaluation},
  author = {Or Honovich and Roee Aharoni and Jonathan Herzig and Hagai Taitelbaum and Doron Kukliansy and Vered Cohen and Thomas Scialom and Idan Szpektor and Avinatan Hassidim and Yossi Matias},
  year = {2022},
  eprint = {2204.04991},
  archivePrefix = {arXiv},
  primaryClass = {cs.CL},
  url = {https://arxiv.org/abs/2204.04991}
}

@misc{dubanowska2025representationbased,
  title = {Representation-based Broad Hallucination Detectors Fail to Generalize Out of Distribution},
  author = {Zuzanna Dubanowska and Maciej {\.Z}elaszczyk and Micha{\l} Brzozowski and Paolo Mandica and Micha{\l} Karpowicz},
  year = {2025},
  eprint = {2509.19372},
  archivePrefix = {arXiv},
  primaryClass = {cs.LG},
  url = {https://arxiv.org/abs/2509.19372}
}

@misc{dassen2026factum,
  title = {FACTUM: Mechanistic Detection of Citation Hallucination in Long-Form RAG},
  author = {Maxime Dassen and Rebecca Kotula and Kenton Murray and Andrew Yates and Dawn Lawrie and Efsun Kayi and James Mayfield and Kevin Duh},
  year = {2026},
  eprint = {2601.05866},
  archivePrefix = {arXiv},
  primaryClass = {cs.CL},
  url = {https://arxiv.org/abs/2601.05866}
}

@misc{zhao2025qwen3guard,
  title = {Qwen3Guard Technical Report},
  author = {Haiquan Zhao and Chenhan Yuan and Fei Huang and Xiaomeng Hu and Yichang Zhang and An Yang and Bowen Yu and Dayiheng Liu and Jingren Zhou and Junyang Lin and others},
  year = {2025},
  eprint = {2510.14276},
  archivePrefix = {arXiv},
  primaryClass = {cs.CL},
  url = {https://arxiv.org/abs/2510.14276}
}

@misc{padhi2024granite,
  title = {Granite Guardian},
  author = {Inkit Padhi and Manish Nagireddy and Giandomenico Cornacchia and Subhajit Chaudhury and Tejaswini Pedapati and Pierre Dognin and Keerthiram Murugesan and Erik Miehling and others},
  year = {2024},
  eprint = {2412.07724},
  archivePrefix = {arXiv},
  primaryClass = {cs.CL},
  url = {https://arxiv.org/abs/2412.07724}
}

@misc{han2025safeswitch,
  title = {SafeSwitch: Steering Unsafe LLM Behavior via Internal Activation Signals},
  author = {Peixuan Han and Cheng Qian and Xiusi Chen and Yuji Zhang and Heng Ji and Denghui Zhang},
  year = {2025},
  eprint = {2502.01042},
  archivePrefix = {arXiv},
  primaryClass = {cs.LG},
  url = {https://arxiv.org/abs/2502.01042}
}

@misc{sreedhar2024canttalkaboutthis,
  title = {CantTalkAboutThis: Aligning Language Models to Stay on Topic in Dialogues},
  author = {Makesh Narsimhan Sreedhar and Traian Rebedea and Shaona Ghosh and Jiaqi Zeng and Christopher Parisien},
  year = {2024},
  eprint = {2404.03820},
  archivePrefix = {arXiv},
  primaryClass = {cs.CL},
  url = {https://arxiv.org/abs/2404.03820}
}

\end{document}